\documentclass[12pt]{article}
\usepackage[dvips]{color}
\usepackage[]{graphicx}
\usepackage{amsthm,amssymb,amsmath,graphics,times}
\usepackage{graphicx}

\usepackage{hyperref}

\usepackage{mathrsfs}

\newtheorem{theorem}{Theorem}

\theoremstyle{definition}

\theoremstyle{remark}

\newcommand{\eref}[1]{(\ref{#1})}

\newcommand{\pmap}{T}
\newcommand{\hD}{\mathscr D}
\newcommand{\act}{\sigma}

\begin{document}

\centerline{\Large \sffamily Error-free Training for Artificial Neural Network}

\bigskip

\centerline{Bo Deng}

\centerline{Department of Mathematics,
University of Nebraska-Lincoln, Lincoln, NE 68588}

\centerline{{\tt bdeng@math.unl.edu}}


\bigskip
\noindent \textbf{Abstract: Conventional training methods
for artificial neural network (ANN) models never achieve
zero error rate systematically for large data. 
A new training method consists
of three steps: first create an auxiliary data from conventionally
trained parameters which correspond exactly to a global minimum
for the loss function of the cloned data; second create a one-parameter
homotopy (hybrid) of the auxiliary data and the original data; and third
train the model for the hybrid data iteratively from the auxiliary data
end of the homotopy parameter to the original data end while maintaining
the zero-error training rate at every iteration. This continuation
method is guaranteed to converge numerically by a theorem which
converts the ANN training problem into a continuation problem for
fixed points of a parameterized transformation in the training
parameter space to which the Uniform Contraction Mapping Theorem from
dynamical systems applies.}

\bigskip\noindent
{\em Key Words}: Artificial neural networks, stochastic gradient descent,
gradient descent tunnelling, zero-error training rate, uniform contraction theorem

\bigskip

\bigskip

By definition, to train an ANN model is 
to find the global minimum of its loss function with the 100\%  
accuracy rate. The theoretical solution to this problem was 
established by \cite{Cybe89,Horn89} in 1989 for finite 
points classification  
with sufficiently many parameters. For small training data, the 
training problem can be solved by the gradient descent (GD) method. 
But for large data, such as the most popular MNIST benchmark 
data for handwritten digit classification (\cite{Lecu98}),
the problem is not known to be solved systematically. 
Currently, the state of art for the
MNIST problem achieved a 99.87\% positive
rate (PR) with about 1.5 millions parameters (\cite{web23}).   
Here below we describe a method to bring the PR for
supervised ANN models to 100\%.

Let $p=(W,b)$ be the weight parameters and the
biases for an ANN with supervised training (\cite{Niel15}).
Let $q=L(p)$ be the loss function for the
ANN with respect to a given training data set $D$.
Denote by $\bar p$ any local
minimum of $L$ and $p^\ast$ the global minimum with 
the perfect accuracy if exists.
To train the ANN is to find this perfect global minimum
$p^\ast$. Currently training is done by a variety of
implementations of the gradient descent
method (GD) (\cite{Robb51, Rose58}).
Specifically, the basic idea works as follows. Let $\nabla L(p)$
denote the gradient of $L$. Then starting at an initial
guess $p_0$, and for a learning rate parameter $\alpha>0$,
the next update $p$ is given by this iterative formula
\begin{equation}\label{eq_conventional_training}
p_{k+1}=p_k-\alpha\nabla L(p_k)
\end{equation}
for $k=0,1,2,\dots$. So far none of the variations has
found the global minimum $p^\ast$ with the 100\% PR for
MNIST's full 60,000 training data.

\bigskip\noindent
\textbf{Theory of Convergence}
\smallskip

The theoretical basis for our method is based on an
equivalent setting for the conventional training method.
Specifically, to train the ANN from any $p_0$ is to find
the gradient flow path $p(t)$ satisfying the induced
gradient system of equations
\[
\dot p(t)=-\nabla L(p(t)),\ \hbox{ with $p(0)=p_0$ for the initial condition.}
\]
All conventional GD methods are discrete approximations of the
gradient flow $p(t)$. For example, the basic searching
algorithm \eref{eq_conventional_training} above
is the numerical implementation of Euler's method for the 
differential equations. In this equivalent setting, any local
minimum point of the loss function is a stable equilibrium
of the gradient system. The converse is also true.

Specifically, let $\phi_t(p_0)$ denote the solution
operator of the gradient system satisfying the initial
condition
\[
\phi_0(p_0)=p_0.
\]
The solution $p(t)$ to the gradient system with
the IC $p(0)=p_0$ is $p(t)=\phi_t(p_0)$. That is,
$\phi_t:\mathbb R^n\to \mathbb R^n$ defines a
transformation or mapping from $\mathbb R^n$ to
itself for every time $t$. Thus, the subscript $0$
can be dropped and $\phi_t(p)$ can be used to
denote the solution operator, mapping point $p$
to $\phi_t(p)$ after time $t>0$.
In this setting, every local
minimum $\bar p$ of the loss function $L$ is a
locally stable fixed point of the solution
operator $\phi_t(p)$ for every $t\ge 0$
\[
\phi_t(\bar p)=\bar p,\ t\ge 0.
\]
Conversely, every locally stable fixed point
of $\phi_t$ is a local minimum point of $L$.
In addition, a fixed point of $\phi_t$ for one
fixed nonzero $t$, say $t=1$, is a fixed point of
$\phi_t$ for all $t>0$ (\cite{Chow82}).
Thus, we only need to consider
the solution operator at one fixed time, say $t=1$,
$T(p):=\phi_1(p)$. Such a map is called a Poincar\'e map.
The conclusion is, a point $\bar p$ is
a locally stable fixed point of the
Poincar\'e map $\pmap$ if and only if $\bar p$
is a local minimum of the loss function $L$.

So far the theory is for supervised training on one set
of training data. We now consider the case that there is a
set of training sets of data, denoted by $\hD_\lambda$
where $\lambda$ is a parameter from a compact interval,
say $0\le\lambda\le 1$. For each $\lambda\in[0,1]$, the
task is to find the global minimum $p^\ast_\lambda$ for the same
ANN model on training data $\hD_\lambda$ whose corresponding
loss function can be denoted by $L_\lambda$. We can call this
type of training a parameterized training with parameter
$\lambda$ or a parameterized co-training.

\begin{figure}[t]
\centerline{\parbox[b]{3.5in}
{\includegraphics[width=3.5in,height=2.4in]{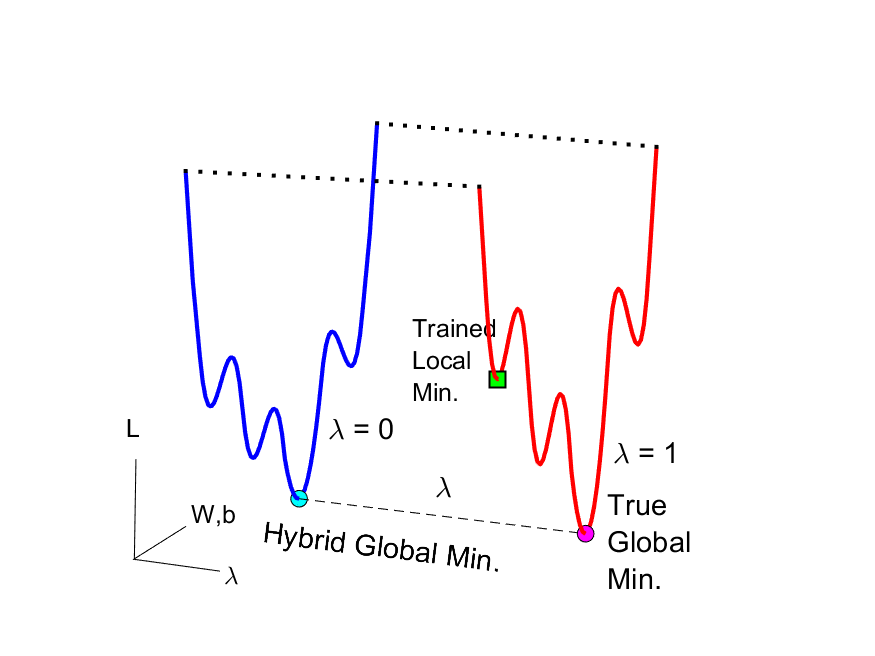}}
\parbox[b]{3.5in}
{\includegraphics[width=3.5in,height=2.4in]{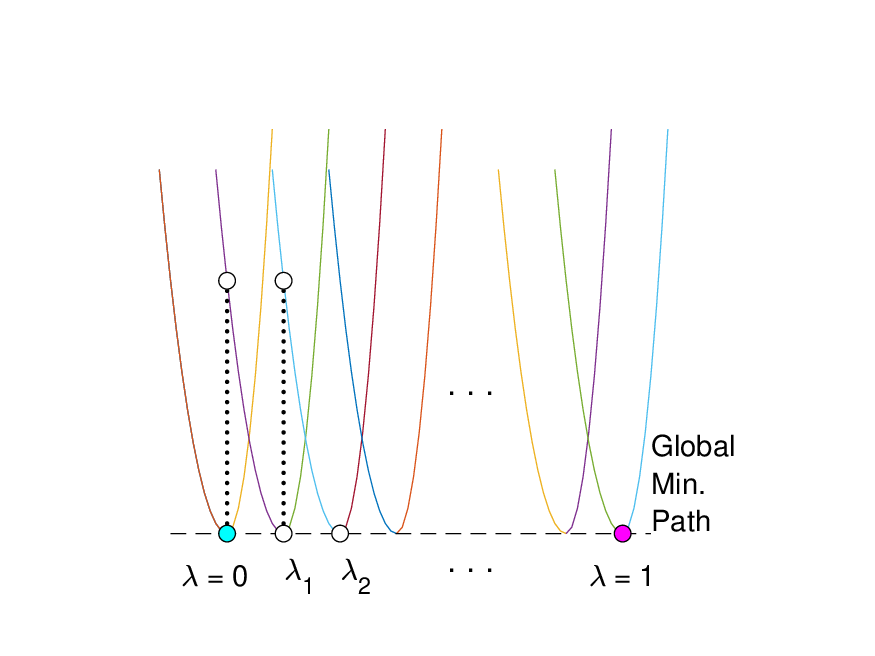}}
} 

\centerline{(a)\hskip 3in (b)}

\caption{\textbf{GM Continuation.}
(a) Trained local minimum
is where a conventional training method ends. The
surface with the co-train parameter $\lambda$
represents the landscape graph for the hybrid loss function.
The method finds the GM below the landscape by
continuing the GM at the $\lambda=0$ end
all the way to the $\lambda=1$ end. (b) The disc on the line and the disc
directly above it are the same parameter point $p$ in
the parameter space $\mathbb R^n$ relative to different
continuation parameter values $\lambda$. The former represents
the GM for a given $\lambda$, seating at the GM for the loss
function. The latter presents the same point lies inside
or outside the basin of attraction, or the potential well,
of the GM for a forwarding co-train parameter value
$\lambda$. If it is inside the GM's potential well, a
gradient descent will lead the point to the new GM. Otherwise,
it will leads it to a local minimum of the $\lambda$
value, in which case backtrack correction is applied
to continue the process. The goal is to maintain the 100\% PR
throughout the continuation from start to finish.}
\label{fig_GDT}
\end{figure}

In terms of the Poincar\'e mapping equivalency, for each
$\lambda$ we have the equivalent Poincar\'e map $\pmap_\lambda$
for which $\bar p_\lambda$ is a local minimum for $L_\lambda$
if and only if $\bar p_\lambda$ is a locally stable fixed
point of $\pmap_\lambda$. Our method is based on the following
theorem.
\begin{theorem}[Continuation Theorem of Global Minimums]
Assume the perfect global minimum point $p^\ast_\lambda\in \mathbb R^n$
of $L_\lambda$ exists for every $\lambda\in [0,1]$ and
is asymptotically stable for the Poincar\'e map
$\pmap_\lambda$. Assume also $\pmap_\lambda$ is continuous
in $\lambda$ and is differentiable at
$p^\ast_\lambda$. Then the global minimums form a continuous
path $\gamma:=\{p^\ast_\lambda:0\le\lambda\le 1\}$ in the
training parameter space $\mathbb R^n$.
\end{theorem}
\begin{proof} For each co-train parameter
$\lambda\in[0,1]$, let $D\pmap_\lambda(p^\ast_\lambda)$
be the linearization of the Poincar\'e map $\pmap_\lambda$
at the fixed point $p^\ast_\lambda$. Since $p^\ast_\lambda$
is asymptotically stable for $\pmap_\lambda$ which is
differentiable, there is an adapted norm (\cite{Deng23})
and a small convex neighborhood $U_\lambda$ of the point so that
$\pmap_\lambda$ is Lipschitz continuous with the Lipschitz
constant smaller than 1, i.e. $\pmap_\lambda$ is locally
contracting. Let $\rho(p)$
be a $C^\infty$ cut-off scalar function
(\cite{Chow82, Deng23}) in $U_\lambda$. Extend
$\pmap_\lambda$ from $U_\lambda$ to the entire space
$\mathbb R^n$ by
\[
\pmap_\lambda(p)\to D\pmap_\lambda(p^\ast_\lambda)p+
\rho(p-p^\ast_\lambda)[\pmap_\lambda(p)-D\pmap_\lambda(p^\ast_\lambda)p]
\]
For small enough neighborhood $U_\lambda$, the
extended map is a contraction mapping in $\mathbb R^n$.
Without loss of generality, we will use the same notation
$\pmap_\lambda$ for the extended map. Because the interval
$[0,1]$ where $\lambda$ is from is compact, the extended
map $\pmap_\lambda$ can be made to be uniformly contracting
for all $\lambda\in[0,1]$. As a consequence by the Uniform
Contraction Mapping Theorem (\cite{Chow82, Deng23}), for
each $\lambda$, $\pmap_\lambda$ has a unique fixed point
which by construction is exactly the global minimum
point $p^\ast_\lambda$ for $L_\lambda$. Also as a
consequence to the Uniform Contraction Mapping Theorem,
because $\pmap_\lambda$ is continuous in $\lambda$,
the set of points $\{p^\ast_\lambda:0\le\lambda\le 1\}$
form a continuous path in the training
parameter space $\mathbb R^n$ that is parameterized
by the co-train parameter $\lambda$.
\end{proof}

\bigskip\noindent
\textbf{Method of Global Minimum Continuation}
\begin{enumerate}
\leftskip -.15in %
\item Let $D$ denote the training
data for an ANN model. For example, for the MNIST benchmark
problem, $D$ is the training data of 60,000 handwritten
digits. Train the model by any conventional ways, e.g. 
stochastic gradient descent (SGD) method, to achieve a
considerable positive rate. This divides the data
set $D$ into those which are correctly labelled or trained,
$D_t$, and those which are incorrectly labelled or untrained,
$D_u$.

\vskip 8pt

\item Create a set of auxiliary data, consisting two parts.
One part is exactly the correctly labelled data $D_t$.
The other part is cloned or duplicated from $D_t$ for
the same number as $D_u$. Specifically, for each incorrectly
labelled data from $D_u$, pick a trained partner data 
from $D_t$, preferably having the same training label and 
without repeating.
Denote this cloned data set by $\bar D_u$. As a result,
the joint data set $\bar D=D_t+\bar D_u$ is
perfectly trained for the ANN model, with 100\% PR,
with the same weight parameter and bias $p=\{W,b\}$ as
for the imperfectly trained but true data $D$.
The corresponding auxiliary system with $\bar D$
is referred to as a training partner. The imperfectly
trained parameters $p$ is automatically the global
minimum, i.e., $p=\bar p^\ast$ for the partner system.

\vskip 8pt

\item Introduce a parameter $\lambda$ in the
interval $[0,1]$. For each $\lambda$, create
an hybrid data from the true data $D$ and the
auxiliary partner data $\bar D$ by
\[
\hD_\lambda=(1-\lambda)\bar D+\lambda D.
\]
This technique by such weighted average of
$\bar D$ and $D$ with weight $(1-\lambda)$
and  $\lambda$, respectively, is known as homotopy
for continuation in dynamical systems
(\cite{Chow78,Li87}).
The trained data $D_t$ remains unchanged
for all $0\le\lambda\le 1$. The part corresponding
to the erred data changes from the perfectly trained
partner set $\bar D_u$ at $\lambda=0$ to the original
data set $D_u$ at $\lambda=1$. If the partnering data
are chosen for the same training labels as the
partneree data, then the same training labels can be
used throughout for all co-train
parameters $0\le \lambda\le 1$. More generally, one
can create a homotopy in the classification space
and directly compute the error-rate from the
classifying vectors. See Fig.\ref{fig_GDT}(a).

\vskip 8pt

\item For the ANN model with the hybrid and co-train data
set $\hD_\lambda$ from $\lambda=0$ to $\lambda=1$, the
Continuation Theorem of Global Minimums guarantees that the
error-free global minimum at $\lambda=0$ for $\hD_0=\bar D$
is connected all the way through $0<\lambda< 1$ to the error-free
global minimum at $\lambda=1$ with $\hD_1=D$, which is what
we want for the solution of the training problem.
Continuation of the GMs from $\lambda=0$ to $\lambda=1$
can be done by any continuation method. For example,
it can be done by solving for equilibrium solutions for
the gradient vector field
\[
\nabla L_\lambda(p)=0
\]
by Newton's method or its variations (\cite{Kell03}),
starting at the partner system's
global minimum $p^\ast_0$ at $\lambda=0$.
Here below we describe a new method referred to
as the gradient-descent tunneling (GDT) method by
way of backtrack correction.
\end{enumerate}

\bigskip\noindent
\textbf{Gradient Descent Tunneling: Backtrack Correction}
\begin{enumerate}
\leftskip -.15in %
\item Start at $\lambda=0$ where the global minimum
$p=p^\ast_0$ locates for the partner system with training
data $\hD_0=\bar D$.

\vskip 8pt

\item Move forward in $\lambda$ to a value $\lambda=a>0$.
If $p^\ast_0$ is inside the gradient-descent's
basin of attraction for the GM $p^\ast_a$ of the loss function
$L_a$, then applying a gradient descent algorithm in a few
iterations to find the GM $p^\ast_a$. Fig.\ref{fig_GDT}(b)
illustrates this case if $a=\lambda_1$.

\vskip 8pt

\item If the step size $\lambda=a$ taken if too large,
say, $a=\lambda_2$ as shown in Fig.\ref{fig_GDT}(b), then
the parameter $p=p^\ast_0$ is inside the basin of attraction
of a local minimum point. Any gradient descent search will
miss the GM $p^\ast_a$. When this occurs, the PR indicator
for the search will be less than 100\%.
In this case, the step size
is reduced to a smaller value, say $a/2$, and then
repeat the same step. This procedure is
referred to as {\em backtrack correction}.

\vskip 8pt

\item The Continuation Theorem of Global Minimums
guarantees the convergence of the algorithm.
That is, the number of backtrack correction
required before the PR becomes 100\% is finite. And
the convergence also guarantees the GM $p^\ast_1$ for the
true system with data $D$ can be reached in finite
steps because the extended Poincar\'e map $\pmap_\lambda$
is a uniform contraction on a compact interval
$\lambda\in [0,1]$. In the continuation
path to $\lambda=1$ and PR $=$ 100\%, there are trained
models with sub-100\% but high positive rates.
\end{enumerate}

\noindent
\textbf{Gradient Descent Tunneling: Adaptive Continuation}
\begin{enumerate}
\leftskip -.15in %
\item[] For computational efficiency, we can speed up
the continuation of $\lambda$ by increasing the
step-size $a$ if the
iteration yields a sequence of consecutive
100\% PRs. When combining with the backtrack correction
technique, such a continuation should result in an efficient
algorithm to find the true global minimum $p^\ast_1$ at
$\lambda=1$.
\end{enumerate}

\bigskip\noindent
\textbf{Cumulative Training}
\begin{enumerate}
\leftskip -.15in %
\item[] As one direct
implication of the continuation method, the
trained data $D_t$ can be viewed as what have
been learned by the ANN model and the erred data
$D_u$ can be viewed as what have to be learnt new
data. The co-train continuation method together with
GDT can be used to accomplish such memory retention
learning tasks so that the model is trained to
the new while keeping the old intact.
\end{enumerate}

\begin{figure}[t]
\centerline
{\scalebox{.6}{\includegraphics{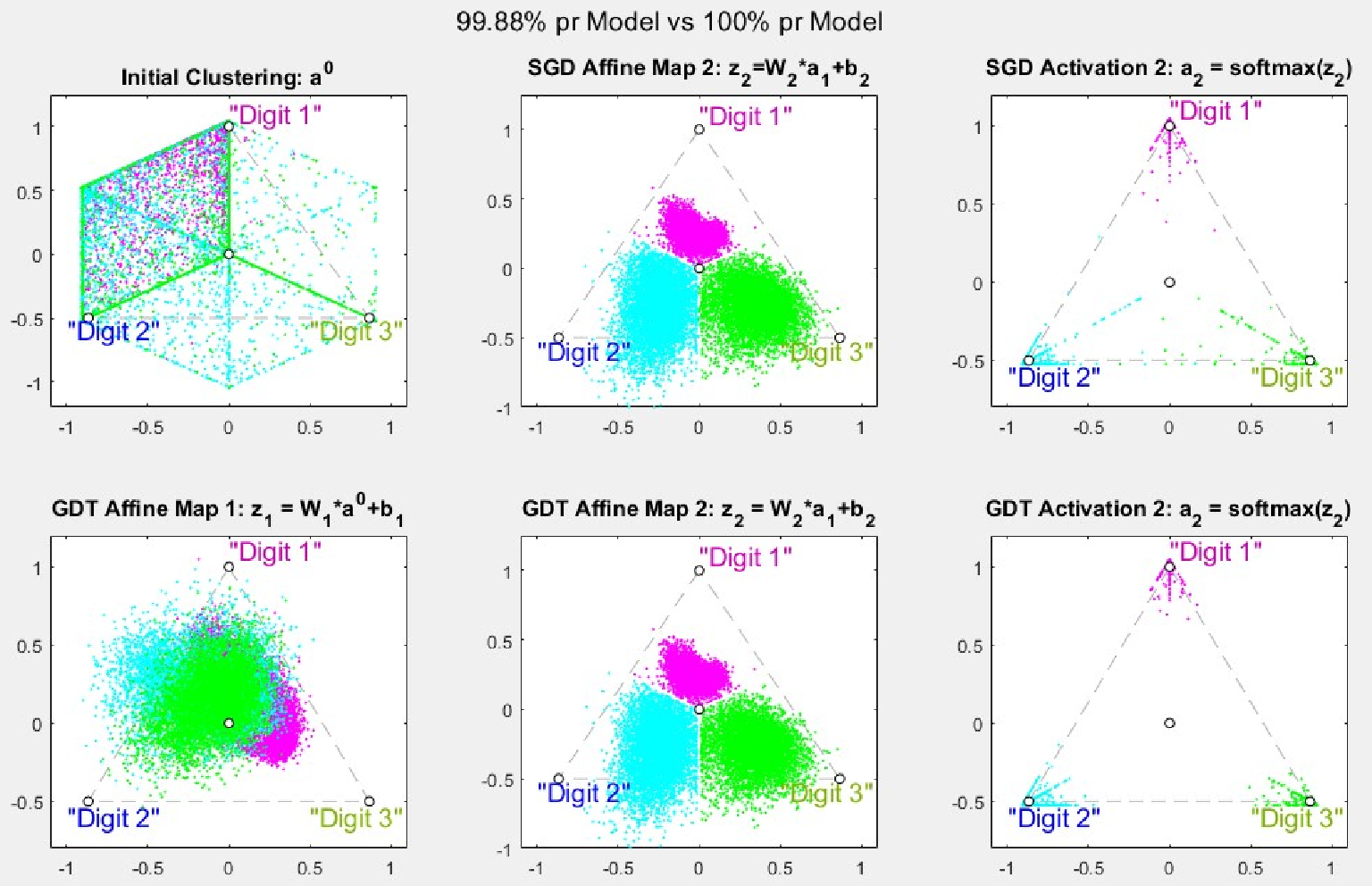}}}
\caption{\textbf{Complete Segmentation.} A projection of the
784-dimension classification space to $\mathbb R^3$, with $[1,0,0],\
[0,1,0],\ [0,0,1]$ for digits ``1", ``2", ``3". The viewing
angle
vector is $[1,1,1]$ towards the origin. Top row is for
an SGD trained model with 99.88\% positive rate and the bottom
row is for a fully trained model. The top-right plot
presents how the current benchmark model would look like
for its final clustering of points. The ANN architecture 
is 784-100-10.}
 \label{fig_cluster}
\end{figure}

\bigskip\noindent
\textbf{Result.} The continuation method with gradient
descent tunneling was implemented on Matlab. Models of two 
different types of activation functions were used, all having 
the 784-$n$-10 architecture, and all achieved 100\% PR 
for the MNIST benchmark data, where $n$ is the number of 
nodes for the only hidden layer. All used  
the softmax for the output layers's classification 
function.

We used the popular ReLU as the activation functions 
for the hidden layer, with node number 
$n=$ 100, 80, 60, 40, 20, respectively. The corresponding numbers of
parameters in $\{W,b\}$ for the 784-$n$-10 models
are 79510, 63610, 47710, 31810, 15910, respectively. 
Fig.\ref{fig_cluster} shows a comparison between
conventionally-trained model and our fully-trained
model. The latter achieves a clean and complete
partition of the 60,000 training points from the
784-dimensional classification space in a 3D
projection for digits 1, 2, and 3. 
Because ReLU is not differentiable at the origin, the 
loss function is not differentiable everywhere as 
required by the continuation theorem. To avoid the 
non-differentiable points, we chose the initial 
parameters in $W$ and $b$ randomly with uniform 
distribution in interval $(-5, 5)$. Instead of 
${\rm softmax}(x)$, we used a scaled version, 
${\rm softmax}(x/100)$ for the output layer's pooling. 
But the continuation 
failed to work for the initials from $(-0.5,0.5)$ 
together with the standard softmax, 
most likely because the continuation can't avoid the 
non-smooth points of the loss function.    

For the second type of activation function, 
we used a new sigmoid function 
\[
\act(x)=\left\{
\begin{array}{ll}
      0, & x\leq 0 \\
      \tanh^2(x/2), & x\geq 0 \\
\end{array}\right.
\]
This function was derived as the voltage-gated 
activation function for neuron models which all modelers 
must agree regardless which approach to take, 
modeling the conductance or modeling the resistance of the 
ion channels (\cite{Deng19}). More specifically, 
the inverse of $\act$ over $x>0$ is
\[
\rho(x)=\coth^2(x/2),\ \  x\geq 0,
\]
having the property that both are solutions to the 
following differential equation 
\[
\frac{dy}{dx}=\sqrt{y}(1-y), 
\]
referred to as the switch equation in \cite{Deng19}, 
and the sigmoid function $y=\act(x)$ is 
refereed to as a {\it switch function} by 
extension, for which 
the state is off if $x\le 0$ and on if $x>0$. 
When the ReLU activation function is replaced by 
the switch function $\act$, the continuation
method works perfectly for initials of $W,b$ from 
$(-0.5,0.5)$ together with the standard  
softmax. Moreover, 
the convergence with the switch activation function 
is at least 3 times faster than with the ReLU model 
for GDT. Also, SGD works faster with the switch 
function than the ReLU, most likely because of the 
smoothness difference between them. 

\begin{figure}[t]
\centerline
{\scalebox{.6}{\includegraphics{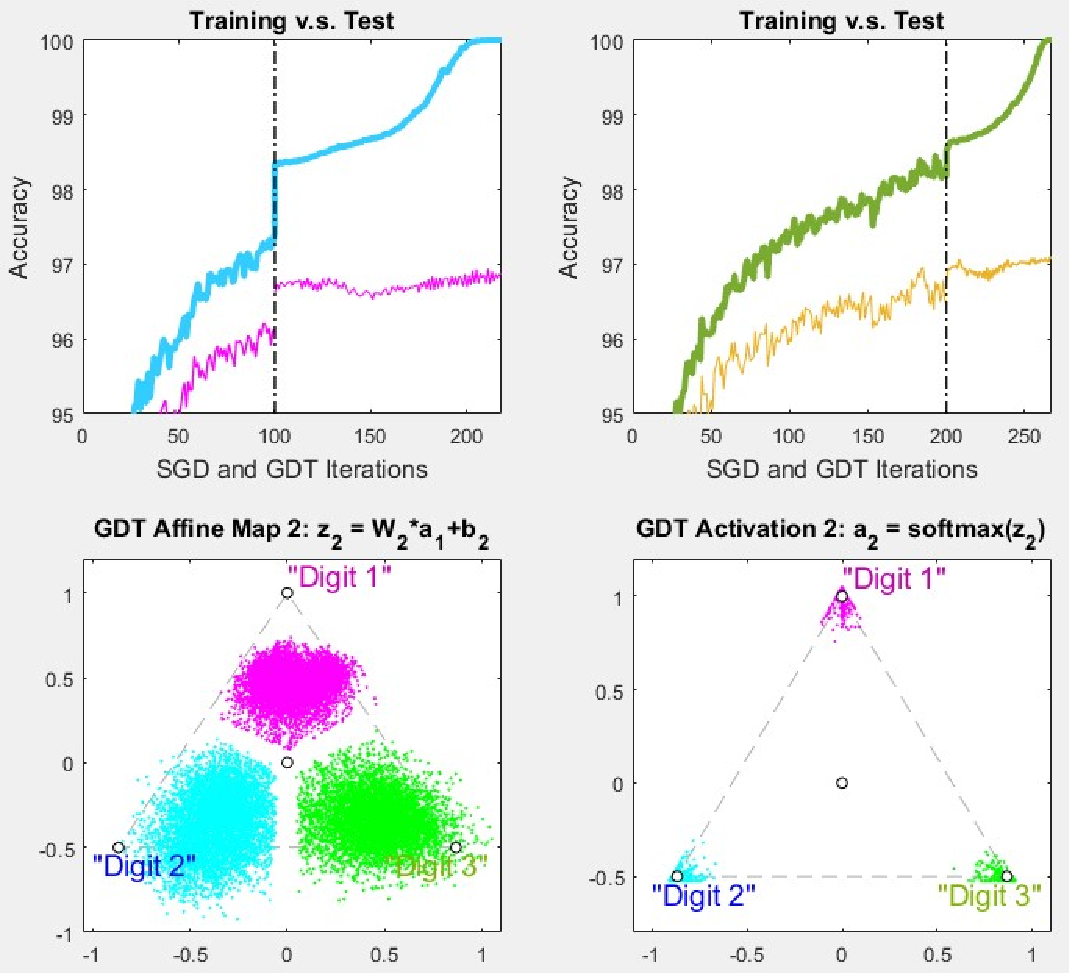}}}
\caption{\textbf{Result with Switch Activation.} Top two plots 
show two trainings for the model with the switch activation 
$\act$ with different number of SGD iterations. Thick lines 
are for the training data and thin lines are for the test 
data. The bottom plots are similar to the clustering plots 
of Fig.\ref{fig_cluster} for a fully-trained model.}
 \label{fig_switch}
\end{figure}

Specifically, Fig.\ref{fig_switch} shows some results for a
784-100-10 model with the switch function for activation. It 
shows the accuracy curves for MNIST's training and test data
by the conventionally trained model with SGD and the 
fully-trained model with GDT. It shows that the final 
accuracy for test after GDT is always better than the 
SGD-trained model. It also shows that 
other than the normal random fluctuations for the test 
accuracy curve, little can be construed as ``overfitting". 
The figure also shows the last and the second last 
segmentations of the transformed training data, with the second 
last achieved a clearer separation than the model with 
ReLU from Fig.\ref{fig_cluster}. 

\bigskip\noindent
\textbf{Discussion.} Mathematical data sciences is to solve 
theoretical and computational scaling problems from small 
data to large data. The Universal Approximation Theorem 
by \cite{Cybe89,Horn89} solved the theoretical problem to 
automate the classification problem by ANN models. 
For small data, the (stochastic) gradient descent method 
is able to solve the training problem, i.e. finding the 
global minimum of a model's loss function. For large 
data, finding the global minimum by SGD becomes a game 
of chance. Our Continuation Theorem should fill 
this computational gap. 

In the parameter $\{W,b\}$ space, there are infinitely many 
global minimums of a model's loss function. It can be 
found by choosing different initial parameters, or by 
different choices of auxiliary training partners, or by 
different training batches or iterations with SGD searching. 
Together with the continuation theorem, it 
suggests that the global minimums form a hyper-surface 
continuum in the parameter space. As a result, 
these global minimums are infinitely many but have  
zero probability to find by chance via SGD when the data 
is big and the dimension of the parameter space is big. 
Our theorem guarantees the convergence of the continuation 
algorithm if there are enough parameters to ensure the 
existence of the hybrid global minimums.  

Our empirical finding on the size of the ANN models which 
solved the MNIST training problem suggests that the minimal 
dimension of the parameter space in which a classification 
problem is solvable may not be too high, i.e., not suffering 
the so-called curse of dimensionality. This is because
every ANN model has its own intrinsic dimension, like 
every physics model having finite many state variables 
for their intrinsic dimensionality. That dimension would fix 
some upper bound in its number of parameters to fully 
determine the model. These two numbers plus the number of 
distinct data whose basins of attraction contain all data points
should define some lower bound for the  minimal number 
of parameters in $\{W,b\}$. Of course, this remains as 
an educated conjecture based on the findings of this paper 
and the theory of dimension analysis (c.f. \cite{Deng18}).  

Our result also suggests the following practices for 
ANN models. The switch function for 
activation is better suited than the ReLU is for 
both SGD training and GDT continuation. Because 
there are infinitely many global minimums, we can 
always find those which are good for test data. 
Thus, we can find one we like 
and then use GDT to incorporate all test data into 
one fully-trained model for deployment. Over-fitting
is never an issue for good model, good math, and 
good code. 
  
The error-free training method
is applicable for all supervised trainings of ANNs,
including convolutional neural networks (CNN), spiking neural
networks (SNN). It is even more advantageous because of its
relatively small numbers of model parameters required,
reducing carbon-footprint for AI training. It can
be implemented on platforms from cloud-frame
supper computers to microchips on mobile devices.
The error-free learning capability of the method will
enable AI to fully enter into many fields
such as inventory logistic, record keeping,
human resource management, fully automated grading
for examinations, precision manufacturing,
health care management, pharmaceutical and medical
expert systems (\cite{Yang23}).
It can build error-free modular systems for search
engines and for large language models which always 
involve supervision in various stages. Our
result may redirect the question of how accurately
an ANN can be trained to how to maximize the
potentials of full-trained models.

\bigskip
\bigskip\noindent
\textbf{\large Declarations}

\small\noindent
\textbf{Ethical approval:} Not Applied.

\small\noindent
\textbf{Competing interests:} None.

\small\noindent
\textbf{Authors' contributions:} Not Applied.

\small\noindent
\textbf{Funding:} None.

\small\noindent
\textbf{Availability of data and materials:} All trained ANN
models mentioned in this article can be downloaded from figshare,
\href{https://doi.org/10.6084/m9.figshare.24328756}{https://doi.org/10.6084/m9.figshare.24328756}
titled `Validation for error-free ANN models on MNIST'.
It also contains Matlab mfiles for the SGD training
and GDT continuation algorithms.

\end{document}